\documentclass[10pt,twocolumn,letterpaper]{article}
\usepackage{iccv}
\usepackage{times}
\usepackage{epsfig}
\usepackage{graphicx}
\usepackage{amsmath}
\usepackage{amssymb}

\usepackage{microtype}
\usepackage{graphicx}
\usepackage{booktabs} 

\usepackage[utf8]{inputenc} 
\usepackage[T1]{fontenc}    
\usepackage{url}
\usepackage{amsfonts}       
\usepackage{nicefrac}       
\usepackage{xcolor}         
\usepackage{bm}
\usepackage{color}
\usepackage{algorithm,algorithmic}
\usepackage{listings,jvlisting}
\usepackage{tabularx}
\usepackage[small,bf]{caption}
\usepackage{wrapfig}
\usepackage{lipsum}
\usepackage{mathtools}
\usepackage{mwe}
\usepackage{multirow}

\usepackage{amsthm}
\theoremstyle{plain}
\newtheorem{theorem}{Theorem}[section]

\theoremstyle{definition}

\theoremstyle{remark}
\newtheorem{proposition}[theorem]{Proposition}
\newtheorem{definition}[theorem]{Definition}

\newtheorem{claim}[theorem]{Claim}

\usepackage[pagebackref=true,breaklinks=true,letterpaper=true,colorlinks,bookmarks=false]{hyperref}

\newcommand{\kld}{D_{\mathrm{KL}}}

\newcommand{\posterior}{p(z|\mathbb{X},\theta)}

\newcommand{\datall}{p(\mathbb{X}|\theta)}
\newcommand{\datadist}{p_{\mathcal{D}}(\mathbb{X})}
\newcommand{\datadistj}{p_{\mathcal{D}}(x_{j})}
\newcommand{\vardist}{q(z|\mathbb{X},\phi)}
\newcommand{\gendist}{p(\mathbb{X}|z,\theta)}
\newcommand{\infdist}{p(z|\mathbb{X}, \theta)}

\newcommand{\objssl}{\mathcal{J}_{\mathrm{SSL}}}
\newcommand{\objcross}{\mathcal{J}_{\mathrm{align}}}
\newcommand{\objuniform}{\mathcal{J}_{\mathrm{uniform}}}
\newcommand{\objkl}{\mathcal{J}_{\mathrm{KL}}}
\newcommand{\argmax}{\mathop{\rm argmax}\limits}

\newcommand{\encoders}{g_{\theta}}
\newcommand{\encoderk}{u_{\theta}}
\newcommand{\encodert}{f_{\phi}}
\newcommand{\elbo}{\mathcal{J}_{\mathrm{ELBO}}}
\newcommand{\posteriorj}{p(z|x_{j}, \theta)}
\newcommand{\gendistj}{p(x_{j}|z, \theta)}
\newcommand{\vardisti}{q(z|x_{i}, \phi)}

\newcommand{\conds}{{j}}
\newcommand{\condt}{{i}}
\newcommand{\softmax}{\operatorname{softmax}}
\newcommand{\cat}{\operatorname{Cat}}

\newcommand{\best}[2]{\underline{\textbf{#1}}$\pm$\scriptsize{#2}}

\newcommand{\third}[2]{#1$\pm$\scriptsize{#2}}

\iccvfinalcopy 

\def\iccvPaperID{9958} 

\ificcvfinal\fi

\begin{document}

\title{Representation Uncertainty in Self-Supervised Learning as Variational Inference}
\author{Hiroki Nakamura\thanks{Equal contribution}\\
Panasonic Holdings Corp.\\
{\tt\small nakamura.hiroki003@jp.panasonic.com}
\and
Masashi Okada\textsuperscript{*}\\
Panasonic Holdings Corp.\\
{\tt\small okada.masashi001@jp.panasonic.com}
\and
Tadahiro Taniguchi\\
Ritsumeikan University\\
{\tt\small taniguchi@em.ci.ritsumei.ac.jp}
}

\maketitle
\ificcvfinal\thispagestyle{empty}\fi


\begin{abstract}
In this study, a novel self-supervised learning (SSL) method is proposed, which considers SSL in terms of variational inference to learn not only representation but also representation uncertainties.
SSL is a method of learning representations without labels by maximizing the similarity between image representations of different augmented views of an image.
Meanwhile, variational autoencoder (VAE) is an unsupervised representation learning method that trains a probabilistic generative model with variational inference.
Both VAE and SSL can learn representations without labels, 
but their relationship has not been investigated in the past.
Herein, the theoretical relationship between SSL and variational inference has been clarified.
Furthermore, a novel method, namely variational inference SimSiam (VI-SimSiam), has been proposed.
VI-SimSiam can predict the representation uncertainty by interpreting SimSiam with variational inference and defining the latent space distribution.
The present experiments qualitatively show that VI-SimSiam could learn uncertainty by comparing input images and predicted uncertainties.
Additionally, we described a relationship between estimated uncertainty and classification accuracy.
\end{abstract}

\section{Introduction}
\begin{figure}[t]
    \centering
    \includegraphics[width=0.99\linewidth]{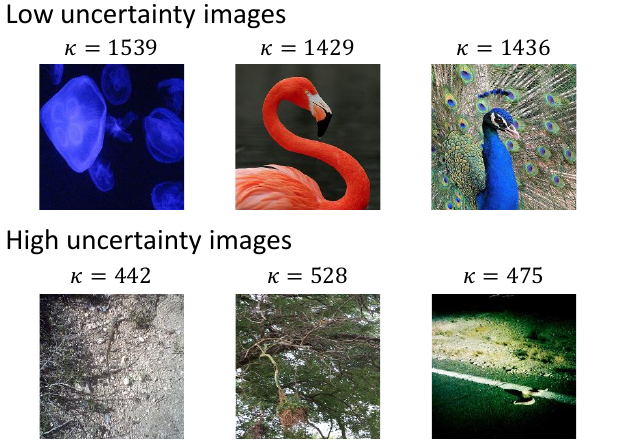}
    \caption{
    \textbf{Input images and the predicted uncertainty parameter $\kappa$.}
    Our method learns not only representations of images but also uncertainties in them without supervision.
    The lower the $\kappa$, the higher the uncertainty.
    The images with a low $\kappa$ appear to have less salient features than those with a high $\kappa$.
    }
    \label{fig:kappa_with_images}
\end{figure}
Self-supervised learning (SSL) is a framework for learning representations of data
~\cite{chen2021exploring, grill2020bootstrap, he2020momentum, chen2021empirical, chen2020simple, zbontar2021barlow, tian2021understanding, newell2020useful, komodakis2018unsupervised, noroozi2016unsupervised}. 
This method enables training of high-performance models in downstream tasks (e.g., image classification and object detection) without substantial manually labeled data through pre-training the network to generate features.
It can mitigate the annotation bottleneck, one of the crucial barriers to the practical application of deep learning.
Some state-of-the-art SSL methods, such as SimSiam~\cite{chen2021exploring}, SimCLR~\cite{chen2020simple}, and DINO~\cite{caron2021emerging}, train image encoders by maximizing the similarity between representations of different augmented views of an image.

The probabilistic generative models with variational inference provide another approach for representation learning~\cite{kingma2013auto}.
This approach learns latent representations in an unsupervised fashion by training inference and generative models (i.e., autoencoders) together.
It can naturally incorporate representation uncertainty by formulating them as probabilistic distribution models (e.g., Gaussian).
However, the pixel-wise objective for reconstruction is sensitive to rare samples~\cite{liu2021self} in such methods.
Furthermore, this representation learning is recently found to be less competitive than the SSL methods on the benchmarking classification tasks~\cite{liu2021self, newell2020useful}.
Although SSL and variational inference seem highly related learning representations without supervision, their theoretical connection has not been fully explored.

In this study, we incorporate the variational inference concept to make the SSL uncertainty-aware and conduct a detailed representation uncertainty analysis.
The contributions of this study are summarized as follows.
\begin{itemize} 
    \item 
    We clarify the relationship between SSL (i.e., SimSiam, SimCLR, and DINO) and variational inference, generalizing the SSL methods as the variational inference of spherical or categorical latent variables (\S\ref{sec:ssl_as_inference}). 
    \item 
    We derive a novel SSL method called variational inference SimSiam (VI-SimSiam) by incorporating the above relationship.
    It learns to predict not only representations but also their uncertainty (\S~\ref{sec:our_method}). 
    %
    \item
    We demonstrate that VI-SimSiam successfully estimates uncertainty without labels
    while achieving competitive classification performance with SimSiam.
    We qualitatively evaluate the uncertainty estimation capability by comparing input images and the estimated uncertainty parameter $\kappa$, as shown in Fig.~\ref{fig:kappa_with_images}.
    In addition, we also describe that the predicted representation uncertainty $\kappa$ is related to the accuracy of the classification task (\S~\ref{sec:experiments}). 
    
\end{itemize} 
A comparison of SimSiam and VI-SimSiam is illustrated in Fig.~\ref{fig:proposed_method}, 
where VI-SimSiam estimates the uncertainty by predicting latent distributions.
%
\section{Related work}
\subsection{Self-supervised learning}
SSL~\cite{chen2020simple, he2020momentum, chen2021exploring, grill2020bootstrap, caron2021emerging} has been demonstrated to have notable performance in many downstream tasks, such as classification and object detection.
Contrastive SSL methods, including SimCLR\cite{chen2020simple} and MoCo~\cite{he2020momentum}, learn to increase the similarity of representation pairs augmented from an image (\textit{positive pairs}) and to decrease the similarity of representation pairs augmented from different images (\textit{negative pairs}).
Conversely, non-contrastive SSL, including SimSiam \cite{chen2021exploring}, BYOL \cite{grill2020bootstrap}, and DINO~\cite{caron2021emerging}, learn a model using only the positive pairs.
Zbontar et al. \cite{zbontar2021barlow} proposed another non-contrastive method using the redundancy-reduction principle of neuroscience.
Furthermore, several studies have theoretically analyzed the SSL methods, such as 
Tian et al. \cite{tian2021understanding} investigated the reasons behind the superior performance of the non-contrastive methods.
Tao et al.~\cite{tao2022exploring} claimed that the (non-)contrastive SSL methods can be unified into one form using gradient analysis.
Zhang~\cite{zhang2022mask} demonstrated a theoretical connection between masked autoencoder~\cite{he2022masked} and contrastive learning.
However, these studies assumed a deterministic formulation without considering uncertainty in the representations.

%
%
\begin{figure}[t]
    \centering
    \includegraphics[width=0.8\linewidth]{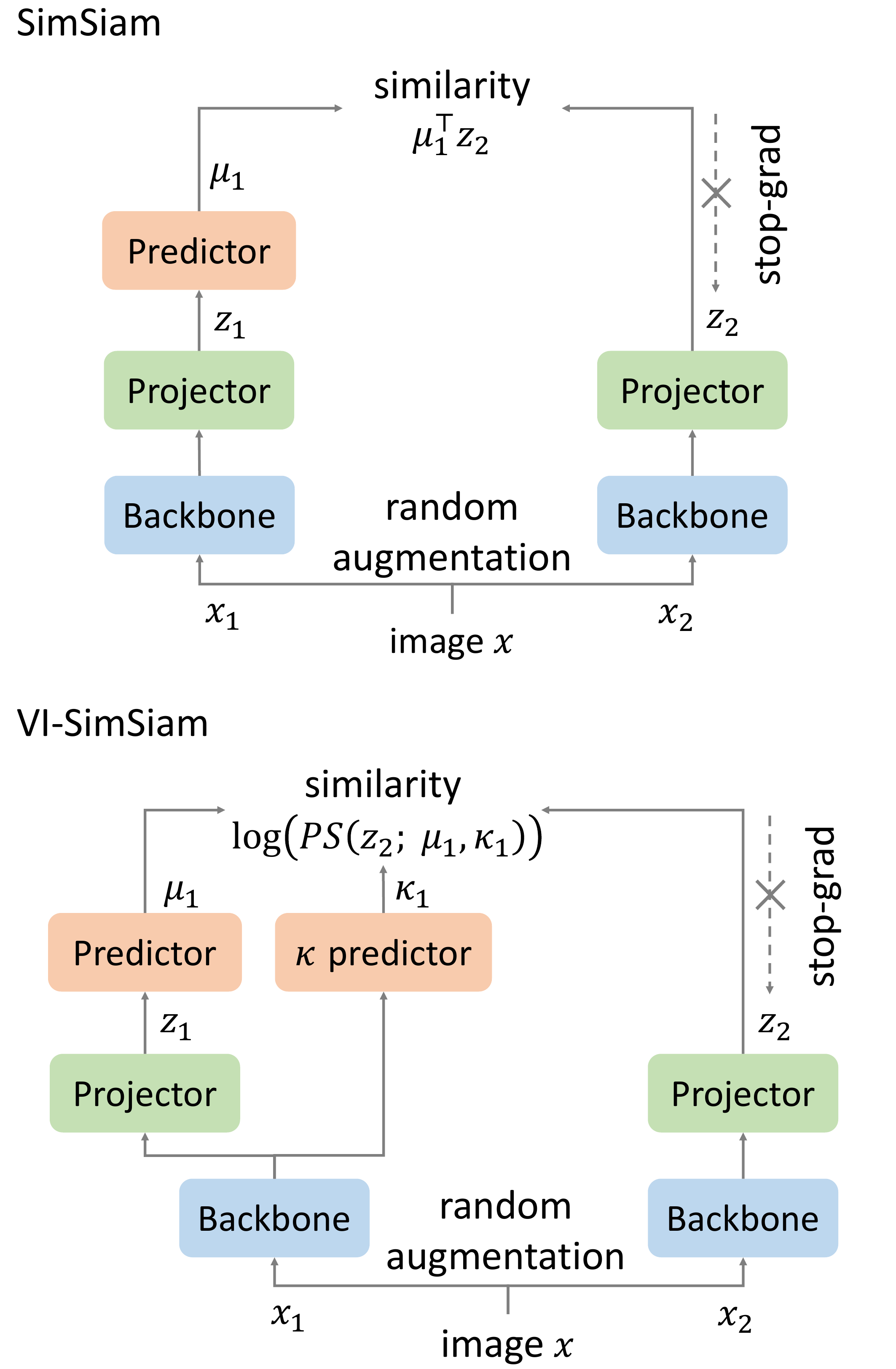}
    \caption{
    Schematics of VI-SimSiam and SimSiam \cite{chen2021exploring}. 
    These methods use the augmented images $x_1$ and $x_2$ as input images.
    In VI-SimSiam, we define the distribution of latent space that follows the Power Spherical distribution \cite{decao2020power}.
    Unlike SimSiam, VI-SimSiam can predict the representation uncertainty of the image $\kappa$.
    %
    }
    \label{fig:proposed_method}
\end{figure}
\subsection{Variational inference}
In deep learning, variational inference is generally formulated using auto-encoding variational Bayes~\cite{kingma2013auto}. The variational inference estimates latent distributions, such as Gaussian~\cite{kingma2013auto}, Gaussian mixture~\cite{tomczak2018vae}, and von Mises-Fischer (vMF) distribution for \textit{hyperspherical} latent space~\cite{davidson2018hyperspherical}.
Although Wang et al.~\cite{wang2020understanding} pointed out that SSL methods learn representations on the \textit{hypersphere}, their relevance to the spherical variational inference~\cite{davidson2018hyperspherical} has yet to be investigated extensively.

A multimodal variational autoencoder~\cite{kurle2019multi, wu2018multimodal, shi2019variational,  sutter2021generalized} is trained to infer latent variables from \textit{multiple observations}\footnote{We treat multiple observations as \textit{multiview} and \textit{multimodal}.} from different modalities.
The latent variable distribution of multimodal variational inference is often assumed to be the \textit{product of experts} or the \textit{mixture of experts} of unimodal distributions \cite{kurle2019multi, wu2018multimodal, shi2019variational}. 
Sutter et al.~\cite{sutter2021generalized} also clarified the connection between them and generalized them as \textit{mixture-of-products-of-experts} (MoPoE) VAE.
Multimodal variational inference seems highly related to the SSL utilizing the \textit{multiview inputs}.
However, their relationship has been unclear.

%
\subsection{Uncertainty-aware methods}
Uncertainty-aware methods \cite{der2009aleatory, kendall2017uncertainties, kendall2018multi, mohseni2020self, winkens2020contrastive} have been proposed to solve the problem of learning hindered by data with high uncertainty.
Kendall and Gal~\cite{kendall2017uncertainties} proposed a method that estimated data uncertainty in regression and classification tasks by assuming that outputs follow a normal distribution.
Scott et al.~\cite{scott2021mises} proposed a stochastic spherical loss for classification tasks based on the von Mises–Fisher distributions.
Additionally, Mohseni et al.~\cite{mohseni2020self} and Winkens et al.~\cite{winkens2020contrastive} presented methods for estimating uncertainty by combining SSL with a supervised classification task.
Uncertainty-aware methods have been proposed for other tasks as well, such as human pose estimation~\cite{petrov2018deep, okada2020multi, gundavarapu2019structured}, optical flow estimation~\cite{ilg2018uncertainty}, object detection~\cite{he2019bounding, hall2020probabilistic}, and reinforcement learning~\cite{lutjens2019safe, okada2020planet, guo2021safety}.

Several studies have suggested methods to incorporate uncertainty in self-supervised learning of specific tasks~\cite{poggi2020uncertainty, pang2020self, xu2021digging, wang2020uncertainty}.
Poggi et al.~\cite{poggi2020uncertainty} proposed an uncertainty-aware and self-supervised depth estimation.
They considered the variance in depth estimated from multiple models as uncertainty.
Wang et al.~\cite{wang2020uncertainty} demonstrated an uncertainty-aware SSL for three-dimensional object tracking, 
wherein the ratio of the distances between positive and negative pairs was treated as uncertainty.
However, these methods hardly discussed representation uncertainty.

\section{Preliminary}\label{preliminary}
The formulations of the SSL methods and variational inference are briefly reviewed in this section.

\subsection{Self-supervised learning methods}
\paragraph*{SimSiam} 
SimSiam is a non-contrastive SSL method with an objective function that is defined as follows;
\begin{align}
  \mathcal{J}_{\mathrm{SimSiam}} \coloneqq \encoders(x_{1})^{\mathsf{T}} \encodert(x_{2}) + \encoders(x_{2})^{\mathsf{T}} \encodert(x_{1}), \label{eqn:obj_non_contr}
\end{align}
where $x_{1}$ and $x_2$ are two augmented views of a single image.
The term $\encoders$ and $\encodert$ are encoders parameterized with $\theta$ and $\phi$, respectively, 
and they map the image $x$ to a spherical latent $z \in \mathbb{S}^{d-1}$.
In the literature on non-contrastive SSL, the two encoders $\encoders$ and $\encodert$ are referred to as \textit{online} and \textit{target} networks, respectively.
SimSiam defines the online network $\encoders$ as $\encoders = h_{\theta} \circ f_{\theta}$, where $h_{\theta}$ and $f_{\theta}$ are referred to as a \textit{predictor network} and \textit{projector network}~\cite{grill2020bootstrap,chen2021exploring}, respectively.

\paragraph*{SimCLR}
The objective function of a contrastive SSL, such as SimCLR is generally described as follows;
\begin{align}
  \mathcal{J}_{\textrm{SimCLR}} &\coloneqq \mathcal{J}^{(1,2)}_{\textrm{SimCLR}} + \mathcal{J}^{(2,1)}_{\textrm{SimCLR}}, \nonumber \\
  \mathcal{J}^{(i,j)}_{\textrm{SimCLR}} &\coloneqq \log \frac{\exp(\encoders(x_{i})^{\mathsf{T}} \encodert(x_{j}))}{\sum_{x' \in \mathcal{B}} \exp(\encoders(x')^{\mathsf{T}} \encodert(x_{j}))}. \label{eqn:obj_contr}
\end{align}
where $\encoders(x), \encodert(x) \in \mathbb{S}^{d-1}$, and $\mathcal{B}$ denotes a minibatch.

%
%
\paragraph*{DINO}
DINO is another non-contrastive SSL with an objective and a latent space (i.e., categorical latent) different from those of SimSiam. 
It is described as;
\begin{align}
  \mathcal{J}_{\textrm{DINO}} &\coloneqq - \mathcal{H}(P_{1,\phi}, P_{2,\theta}) - \mathcal{H}(P_{2,\phi}, P_{1,\theta}) \label{eqn:dino_objective}, \\
  P_{i,\phi} &\coloneqq \softmax\left(({\encodert(x_{i}) - c})/{\tau_{\phi}}\right), \label{eqn:dino_pi} \\
  P_{j,\theta} &\coloneqq \softmax\left({\encoders(x_{j})}/{\tau_{\theta}}\right), \label{eqn:dino_pj}
\end{align}
where $\mathcal{H}(P, Q) (= \sum P \log Q)$ denotes cross entropy between two probabilities, $\encoders(x), \encodert(x) \in \mathbb{R}^{d}$, and ($\tau$, $c$) are the parameters for $sharpening$ and $centering$ operations discussed later in \S~\ref{sec:dino}.


\subsection{Multimodal generative model and inference}
\begin{wrapfigure}{t}[0pt]{0.2\textwidth}
  \centering
  \includegraphics[width=0.175\textwidth]{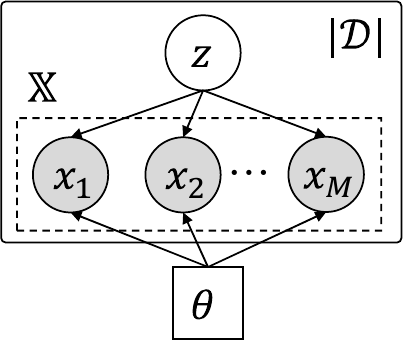}
  \caption{
  Graphical model.
  }
  \label{fig:graphical_model}
\end{wrapfigure}
Fig.~\ref{fig:graphical_model} shows a graphical model for multimodal generative models, where
$\mathcal{D}$ indicates a dataset,
$\mathbb{X}=\{x_{i}\}^{M}_{i=1}$ is a set of multimodal observations $x_{i}$, $z$ is a latent variable of the observations,
$\theta$ is a deterministic parameter of the generative model $\gendist$,
and $M$ is the number of modalities corresponding to the data augmentation types in this paper\footnotemark[1]. 
In the SSL context, $\mathbb{X}$ can be regarded as augmented images from stochastic generative processes.

The objective is to find a parameter $\theta^{*}$ that maximizes marginal observation likelihood;
\begin{align}
  \theta^{*} &= \argmax_{\theta} \mathcal{J} = \argmax_{\theta} \mathbb{E}_{p(z)}[\log \gendist]. \label{eqn:gen_model}
\end{align}
Since the marginalization $\mathbb{E}_{p(z)}[\cdot]$ is intractable, we can instead maximize the evidence lower bound (ELBO);
\begin{align}
  & \mathbb{E}_{p(z)}[\log \gendist] \geq \elbo \coloneqq \nonumber \\
  & = \mathbb{E}_{\vardist}\left[\log \gendist \right] - \kld[{\vardist|p(z)}], \label{eqn:elbo}
\end{align}
where $\vardist$ is a variational inference model parameterized with $\phi$.
By optimizing $\elbo$ with respect to both $\theta$ and $\phi$,
$\vardist$ approaches the posterior $\posterior$ as the following relation holds;
\begin{align}
  \mathcal{J} - \mathcal{J}_{\mathrm{ELBO}} = \kld[\vardist | \infdist] \geq 0,
\end{align}
where $\kld[\cdot]$ is the Kullback-Leibler divergence.
Notably, the posterior varies during the optimization process since it depends on the parameterized generative model; i.e., $\posterior \propto \gendist p(z)$.

\section{Self-supervised learning as inference}\label{sec:ssl_as_inference}
This section shows a connection between SSL and multimodal variational inference. 
Usually, a generative model $\gendist$ and variational inference model $\vardist$ are realized with deep neural networks to solve the problem of Eq.~(\ref{eqn:gen_model}), and
they are trained via ELBO optimization.
Instead, let us consider directly realizing the posterior $\posterior$ as deep neural networks.
For this purpose, we remove the generative model term in Eq.~(\ref{eqn:elbo}) by applying Bayes' theorem;
\begin{align}
  \gendist = \infdist \datall / \mathbb{E}_{\datall}[\infdist]. \label{eqn:gen_approx}
\end{align}
Since $\datall$ is intractable, we approximate it with the empirical data distribution $\datadist$.
Substituting Eq.~(\ref{eqn:gen_approx}) into Eq.~(\ref{eqn:elbo}) and applying the approximation yields a new objective;
\begin{align}
  \objssl &\coloneqq \objcross + \objuniform + \objkl + \datadist, \nonumber \\
  & \stackrel{+}{=} \objcross + \objuniform + \objkl, \label{eqn:ssl_obj}
\end{align}
where,
\begin{align}
  & \objcross \coloneqq \mathbb{E}_{\vardist}\left[\log \posterior \right], \label{eqn:obj_cross} \\
  & \objuniform \coloneqq \mathbb{E}_{\vardist}\left[- \log p_{\mathcal{D}}(z|\theta) \right], \label{eqn:obj_uniform}\\
  & \objkl \coloneqq - \kld[{\vardist | p(z)}], \\
  & p_{\mathcal{D}}(z|\theta) \coloneqq \mathbb{E}_{\datadist}[\infdist].
\end{align}
Furthermore, let $\posterior$ and $\vardist$ respectively be Product-of-Experts (PoE) and Mixture-of-Experts (MoE) of the single modal inference models;
\begin{align}
  \posterior &= \eta_{\theta} \prod^{M}_{j=1} \posteriorj, \label{eqn:poe_posterior} \\
  \vardist &\coloneqq \frac{1}{M} \sum^{M}_{i=1} \vardisti, \label{eqn:moe_variational}
\end{align}
where $\eta_{\theta}^{-1} \coloneqq \int \prod \posteriorj dz$ is the renormalization term. 
Then, we can rewrite $\objcross$ as a form that encourages aligning latent variables from different modals;
\begin{align}
  \objcross =& \sum_{i,j} \mathbb{E}_{\vardisti}\left[\log \posteriorj\right] + M \log \eta_{\theta},
  \label{eqn:obj_cross2}
\end{align}
Eq.~(\ref{eqn:poe_posterior}) is from Prop.~\ref{prop:poe} described below.
Eq.~(\ref{eqn:moe_variational}) is the definition theoretically validated in \cite{shi2019variational,sutter2021generalized}.
Practically, we can ignore unimodal comparisons (i.e., $i=j$) since they provide less effective information.
\proposition \label{prop:poe}
Let $p(z)$ be a non-informative prior.
The multi-modal posterior $\posterior$ takes the form of PoE of the single-modal posteriors $\posteriorj$.
\proof See Appx.~\ref{appx:proof_poe}.

We claim that Eq.~(\ref{eqn:ssl_obj}) generalizes the objectives in Eqs.~(\ref{eqn:obj_non_contr}), (\ref{eqn:obj_contr}) and (\ref{eqn:dino_objective})
as summarized in Table~\ref{tab:ssl_generalization}.
In the rest of this section, we describe how to recover the objectives in the table from Eq.~(10).
In the derivations, the term $M \log \eta_{\theta}$ is ignored by approximating it as a constant (denoted as $\stackrel{+}{\simeq}$).

\begin{table*}
  \centering
  \caption{The summary of SSL methods interpreted as variational inference.} \label{tab:ssl_generalization}
  \footnotesize
  \begin{tabular}{c|cc|c||c}
    \toprule
    & SimSiam~\cite{chen2021exploring} (\S\ref{sec:byol_simsiam}) & SimCLR~\cite{chen2020simple} (\S\ref{sec:simclr_moco}) & DINO~\cite{tao2022exploring} (\S\ref{sec:dino}) & VI-SimSiam (ours) (\S\ref{sec:our_method})\\
    \midrule
    $p(z)$ & \multicolumn{2}{c|}{$\mathcal{U}(\mathbb{S}^{d-1})$} & $\mathcal{U}(\Delta^{d-1})$ & $\mathcal{U}(\mathbb{S}^{d-1})$ \\
    $\vardisti$ & \multicolumn{2}{c|}{Deterministic} & Categorical & Deterministic \\
    $\posteriorj$ & \multicolumn{2}{c|}{von-Mises-Fisher} & Categorical & Power Spherical \\
    \midrule
    $\objcross$
      & $z_{\conds}^{\mathsf{T}} z_{\condt}$ 
      & \multirow{2}{*}{$\log \frac{\exp({z_{\conds}^{\mathsf{T}}z_{\condt}})}{\sum_{z} \exp({z^{\mathsf{T}}z_{\condt}})}$}
      & $\sum P_{j} \cdot \log P_{i}$
      & $\log C_{\mathrm{PS}}(\kappa_{\conds}) +  \kappa_{\conds} \operatorname{log1p}(z_{j}^{\mathsf{T}}z_{i})$ \\
    $\objuniform$
      & DirectPred
      &
      & Centering
      & DirectPred \\
    $\objkl$
      & \multicolumn{2}{c|}{Const.}
      & Sharpening
      & Const. \\
    \midrule
    Uncertainty & \multicolumn{2}{c|}{No} & Yes & Yes \\
    aware & \multicolumn{2}{c|}{(uncertainty parameter $\kappa$ is \textit{fixed})} & (but has not been discussed) & (uncertainty parameter $\kappa$ is \textit{estimated}) \\
    \bottomrule
  \end{tabular}
\end{table*}

\subsection{SimSiam as inference} \label{sec:byol_simsiam}
First, we discuss the relationship between SimSiam and the following definition involving the hyperspherical space $\mathbb{S}^{d-1}$.
\definition{ \label{def:simsiam}
\begin{align}
  & p(z) \coloneqq \mathcal{U}(\mathbb{S}^{d-1}), \\
  & \vardist \coloneqq \frac{1}{M}\sum^{M}_{i=1} \delta(z - \encodert(x_{i})), \label{eqn:varpost} \\
  & \posterior \coloneqq \eta_{\theta} \prod^{M}_{j=1} \mathrm{vMF}(z; \mu=\encoders(x_{j}), \kappa), \label{eqn:vmf_poe} \\
  & \mathrm{vMF}(z; \mu, \kappa) \coloneqq C_{\mathrm{vMF}}(\kappa)\exp(\kappa \mu^{\mathsf{T}} z),
\end{align}
}
where $\encoders(x), \encodert(x) \in \mathbb{S}^{d-1}$, and 
$\mathcal{U}(\mathbb{S}^{d-1})$ is the uniform distribution on the hypersphere.
The term $\mathrm{vMF}(z; \mu, \kappa)$ is the von-Mises-Fisher distribution~\cite{mardia2000directional}
parameterized with the mean direction $\mu \in \mathbb{S}^{d-1}$ and concentration (inversed variance) $\kappa \in \mathbb{R}^{+}$.
The term $C_{\mathrm{vMF}}(\kappa)$ is a normalization constant defined with the modified Bessel function.
In \S\ref{sec:ssl_as_inference}, $\kappa$ and $C_{\mathrm{vMF}}(\kappa)$ are defined to be constants.

\paragraph*{Relation to $\objcross$}
\claim{ \label{claim:obj_non_contr}
Under Def.~\ref{def:simsiam}, we can recover $\mathcal{J}_{\mathrm{SimSiam}}$ in Eq.~(\ref{eqn:obj_non_contr}) from $\objcross$;
\begin{align}
  \objcross  \stackrel{+}{\simeq} \sum_{i,j} g^{\mathsf{T}}_{\theta}(x_{j}) \encodert(x_{i}),
  \label{eqn:byol_objective}
\end{align}
}
\proof
See Appx.~\ref{appx:proof_obj_non_contr}.
%

\paragraph*{Relation to $\objuniform$}
\claim{
    The presence of the predictor $h_{\theta}$ implicitly maximizes $\objuniform$.
}

\proof


Here, we borrow theoretical findings from the DirectPred literature~\cite{tian2021understanding}.
In the literature, considering that $\encoders = h_{\theta} \circ f_{\theta}$, the predictor $h_{\theta}$, defined as the following linear model, can lead to successful convergence;
\begin{align}
  h_{\theta}(z) \coloneqq (U\Lambda^{\frac{1}{2}} U^{\mathsf{T}}) z, \label{eqn:direct_pred}
\end{align}
where $U$ and $\Lambda$ are the eigenvectors and eigenvalues of the covariance matrix of random variables $z$ from $f_{\theta}(x)$; i.e.,
\begin{align}
  \mathbb{E}_{p_{\mathcal{D}}(x)}[f_{\theta}(x)f^{\mathsf{T}}_{\theta}(x)] = U \Lambda U^{\mathsf{T}}.  
\end{align}
With Eq.~(\ref{eqn:direct_pred}), the cosine similarity can be rewritten as%
\footnote{
$g^{\mathsf{T}}_{\theta}(x_{j}) f_{\phi}(x_{i}) = ((U\Lambda^{\frac{1}{2}} U^{\mathsf{T}}) z_{j})^{\mathsf{T}} z_{i} = (U^{\mathsf{T}}z_{j})^{\mathsf{T}}\Lambda^{\frac{1}{2}}U^{\mathsf{T}} (U U^{\mathsf{T}}z_{i})$, where $U U^{\mathsf{T}} = I$.
};
\begin{align}
  g^{\mathsf{T}}_{\theta}(x_{j}) f_{\phi}(x_{i}) = (U^{\mathsf{T}}z_{j})^{\mathsf{T}} \Lambda^{\frac{1}{2}} (U^{\mathsf{T}}z_{i}), 
\end{align}
where $z_{j} = f_{\theta}(x_{j})$ and $z_{i} = f_{\phi}(x_{i})$ are projected by $U^{\mathsf{T}}$ (i.e., projection matrix of PCA) so that each dimension is as independent as possible.
In addition, 
the eigenvalues $\Lambda$ converge to the identity matrix, i.e., the variances of $z$ are constant.
These behaviors encourage the marginalized $z$-distribution to be \textit{isotropic}, i.e., $\mathcal{U}
(\mathbb{S}^{d-1})$.
Consequently, $\objuniform$ is maximized since it takes the maximum iff $p_{\mathcal{D}}(z|\theta) = \mathcal{U}(\mathbb{S}^{d-1})$.

\paragraph*{Relation to $\objkl$}
Since $\vardist$ is deterministic, $\objkl$ is constant.



\subsection{SimCLR as inference} \label{sec:simclr_moco}
Further, we derive the contrastive learning objective with Def.~\ref{def:simsiam}.
$\objkl$ is constant,
as in the previous section, and hence, is disregarded in the current section.
%
\paragraph*{Relation to $\objcross$ and $\objuniform$}
Converse to the previous non-contrastive case, we consider explicitly optimizing $\objuniform$, leading to the following claim;
\claim{ \label{claim:obj_contr}
Under Def.~\ref{def:simsiam}, we can recover $\mathcal{J}_{\mathrm{SimCLR}}$ in Eq.~(\ref{eqn:obj_contr}) from $\objcross$ and $\objuniform$;
\begin{align}
  & \objcross + \objuniform \stackrel{+}{\simeq} \nonumber \\
  &  \sum_{i,j} \mathbb{E}_{\vardisti} \left[
    \log
    \frac{\exp(\encoders(x_{i})^{\mathsf{T}} \encodert(x_{j}))}{\sum_{x' \sim \mathcal{B}} \exp(\encoders(x')^{\mathsf{T}} \encodert(x_{j}))}
    \right]. \label{eqn:obj_contr2}
\end{align}
}
\proof{
See Appx.~\ref{appx:proof_obj_contr}.
}

Note that as per Ref.~\cite{wang2020understanding}, Eq.~(\ref{eqn:obj_contr}) can be decomposed into two objectives similar to $\objcross$ and $\objuniform$.
In contrast, our derivation here aimed to derive Eq.~(\ref{eqn:obj_contr}) in a general form from variational inference.

\subsection{DINO as inference}\label{sec:dino}
Finally, we derive the objective of DINO with the following definitions, where $z$ is defined in the simplex $\Delta^{d-1}$,
\definition{\label{def:dino}
\begin{align}
  & p(z) \coloneqq \mathcal{U}(\Delta^{d-1}), \\
  & \vardist \coloneqq \frac{1}{M}\sum^{M}_{i=1} \cat\left(z; P_{i,\phi}\right), \label{eqn:vardist_dino} \\
  & \posterior \coloneqq \eta_{\theta} \prod^{M}_{j=1} \cat\left(z; P_{j,\theta}\right), \label{eqn:post_dino}
\end{align}
}
where the definitions of $P_{i,\phi}$ and $P_{j,\theta}$ follow Eqs.~(\ref{eqn:dino_pi}) and (\ref{eqn:dino_pj}), 
and $\cat$ means categorical distribution.

\paragraph*{Relation to $\objcross$}
\claim{ \label{claim:obj_dino}
Under Def.~\ref{def:dino}, we can recover $\mathcal{J}_{\mathrm{DINO}}$ in Eq.~(\ref{eqn:dino_objective}) from $\objcross$;

%
\begin{align}
  \objcross \stackrel{+}{\simeq} \sum_{i,j} - \mathcal{H}(P_{i,\phi}, P_{j,\theta}), \label{eqn:obj_dino_derived}
\end{align}
}
\proof{
Substituting Eqs.~(\ref{eqn:vardist_dino}) and (\ref{eqn:post_dino}) into Eq.~(\ref{eqn:obj_cross}) derives Eq.~(\ref{eqn:obj_dino_derived}).
}

\paragraph{Relation to $\objuniform$}
\claim{
  \textit{Centering} in Eq.~(\ref{eqn:dino_pi}) optimizes $\objuniform$.
}

\proof
The centering parameter $c$ in Eq.~(\ref{eqn:dino_pi}) is determined and updated with an exponential moving average of batch mean of $f_{\phi}$; $c \leftarrow mc + (1-m) \mathbb{E}_{\mathcal{B}}[f_{\phi}(x)]$~\cite{caron2021emerging}, where $m$ is a rate parameter.
This batch-norm-like centering encourages the marginalized distribution $q(z|\phi) = \mathbb{E}_{p_\mathcal{D}(x)}[q(z|x, \phi)]$ to be $\mathcal{U}(\Delta^{d-1})$%
\footnote{
In contrast, batch-norm makes the batch distribution $\mathcal{N}(0, 1)$~\cite{ioffe2015batch}.
}
As a result, $p_{\mathcal{D}}(z|\theta)$ will also be uniform while optimizing $\objcross$, thus maximizing $\objuniform$.

\paragraph{Relation to $\objkl$}
\claim{
  \textit{Sharpening} in Eq.~(\ref{eqn:dino_pi}) regularizes $\objkl$.
}
\proof
Since $p(z)$ is defined to be uniform, $\objkl$ acts as an entropy regularizer to the variational distribution $\vardist$.
With a carefully designed temperature $\tau_{\phi}$, sharpening directly regularizes the entropy for learning success; e.g., $\tau_{\phi}$ must be within 0.02 and 0.06~\cite{caron2021emerging}.

\subsection{Discussion}\label{sec:discussion}
\paragraph{Relation to another SSL method}
Moreover, we can integrate Barlow Twins~\cite{zbontar2021barlow} into this \textit{SSL as inference} framework.
Barlow Twins considers the cross-correlation matrix $\mathbb{E}_{\mathcal{B}}[\encoders(x_{i})\encodert(x_{j})^{\mathsf{T}}]$ and optimizes its diagonal and non-diagonal elements to be ones and zeros, corresponding to the optimization of $\objcross$ and $\objuniform$, respectively.

\paragraph{Limitations}
In the derivation, we made some assumptions; $\datall \simeq \datadist$ in Eq.~(\ref{eqn:gen_approx}), $\vardist$ is MoE (Eq.~(\ref{eqn:moe_variational})), and the renormalization term $\eta_{\theta}$ can be ignored in Eq.~(\ref{eqn:obj_cross2}).
The $\objssl$ derived under these assumptions is not guaranteed to be the lower bound of $\mathcal{J}$ in Eq.~(\ref{eqn:gen_model}).
However, the previous success of the SSL method empirically indicates that the assumptions are valid and the optimization progressed while keeping $\mathcal{J} \geq \objssl$.
We hypothesize that to update the target network $f_{\theta}$ such as stop gradient and exponential moving average (EMA)~\cite{grill2020bootstrap,chen2021exploring,caron2021emerging,halvagal2022predictor}, the heuristics may contribute to the above optimization behavior.
However, further theoretical analysis is necessary for future work.
Notably, the stop gradient and EMA can be interpreted as the EM algorithm in variational inference~\cite{bishop2006pattern}.
It is also an attractive research direction to extend SSL methods by relaxing the assumption of $\vardist$; e.g., changing its form from MoE to MoPoE~\cite{sutter2021generalized}, and replacing the deterministic $\delta$ in Eq.~(\ref{eqn:varpost}) with probability distributions.

\paragraph{Representation Uncertainty}
As shown in Table~\ref{tab:ssl_generalization}.
DINO inherently can estimate uncertainty because it fully estimates categorical parameters. 
However, its ability has hardly been discussed.
Conversely, although SimSiam and SimCLR assumed hyperspherical distributions,
the uncertainty parameter $\kappa$ is fixed, thus missing the ability for uncertainty estimation.
To bridge the gap of uncertainty awareness, we propose a new uncertainty-aware method VI-SimSiam  by extending SimSiam in \S\ref{sec:our_method}.
The capability of uncertainty estimation of VI-SimSiam and DINO is subsequently evaluated in \S\ref{sec:experiments}.

\section{Variational inference SimSiam}\label{sec:our_method}
This section introduces a novel uncertainty-aware SSL method by extending SimSiam based on the \textit{SSL as inference} principle.
The previous derivation assumes that $\kappa$ is constant. 
We relax this assumption and allowed $\kappa$ to be estimated by a newly introduced encoder $\encoderk: x \rightarrow \kappa$.
Additionally, we replace the vMF distribution with another spherical distribution, i.e., Power Spherical (PS) distribution~\cite{decao2020power}.
This is because estimating gradients through the modified Bessel function in $C_\mathrm{vMF}(\kappa)$ is computationally expensive and unstable~\cite{decao2020power}.
The modified posterior is defined as:
\begin{align}
  & \posterior \coloneqq \eta_{\theta}\prod^{M}_{j=1} \mathrm{PS}(z; \mu=\encoders(x_{j}), \kappa=\encoderk(x_{j})) \label{eqn:new_posterior},  \\
  & \mathrm{PS}(z; \mu, \kappa) \coloneqq C_{\mathrm{PS}}(\kappa)(1 + \mu^{\mathsf{T}}z)^{\kappa}. 
  \label{eqn:psd}
\end{align}
The normalization constant of the PS distribution $C_{\mathrm{PS}}(\kappa)$ is defined with the beta distribution.
It can be efficiently computed using general deep-learning frameworks.
Substituting Eqs.~(\ref{eqn:varpost}) and (\ref{eqn:new_posterior}) into Eq.~(\ref{eqn:obj_cross}) yields:
\begin{align}
    & \objcross \stackrel{+}{\simeq} \label{eqn:psd_loss} \\
    & \sum_{i,j} \left( \log C_{\mathrm{PS}} \left( \encoderk(x_{j}) \right)
        + \encoderk(x_{j}) \operatorname{log1p} \left(\encoders(x_{j})^{\mathsf{T}} f_{\phi}(x_i) \right) \right), \nonumber
\end{align}
where $\operatorname{log1p}(x) \coloneqq \log(1+x)$.
Similar to Eqs.~(\ref{eqn:obj_non_contr}) and (\ref{eqn:byol_objective}), this loss maximizes the cosine similarity of features from different models, but the similarity term is weighted by $\kappa$.

A novel method referred to as variational inference SimSiam (VI-SimSiam) that optimizes Eq.~(\ref{eqn:psd_loss}) is proposed.
The architecture and pseudo code are shown in Fig.~\ref{fig:proposed_method} and Alg.~{\ref{alg:code}}, respectively, where the number of modalities is $M=2$ and the single-modal comparisons are ignored.
%
\begin{figure}[t]
\centering
\begin{minipage}{.99\linewidth}
\begin{algorithm}[H]
\caption{Pseudocode of VI-SimSiam, PyTorch-like}
\label{alg:code}
\definecolor{codeblue}{rgb}{0.25,0.5,0.5}
\definecolor{codekw}{rgb}{0.85, 0.18, 0.50}
\lstset{
  inputencoding=utf8,
  backgroundcolor=\color{white},
  basicstyle=\fontsize{7.5pt}{7.5pt}\ttfamily\selectfont,
  columns=fullflexible,
  breaklines=true,
  captionpos=b,
  commentstyle=\fontsize{7.5pt}{7.5pt}\color{codeblue},
  keywordstyle=\fontsize{7.5pt}{7.5pt}\color{codekw},
}
\begin{lstlisting}[language=python]
# backbone: ResNet-backbone
# f: Projector
# h, u: Predictor for mu and kappa

# Definition of power spherical dist.
class PSd(Distribution):
  def _init_(self, mu, kappa):
     ...

for x in loader:  # Load a minibatch x
  x1, x2 = aug(x), aug(x)  # Apply augmentation
  y1, y2 = backbone(x1), backbone(x2)
  z1, z2 = f(y1), f(y2)  # Project

  mu1, mu2 = h(z1), h(z2) # Predict mu
  kappa1, kappa2 = u(y1), u(y2) # Predict kappa
  p1, p2 = PSd(mu1, kappa1), PSd(mu2, kappa2)

  # Stop gradient & compute loss
  z1, z2 = z1.detach(), z2.detach()
  L = - p1.log_prob(z2) - p2.log_prob(z1) 
  L.mean().backward()  # Back-prop.
  update(backbone, f, h, u)  # SGD update
\end{lstlisting}
\end{algorithm}
\end{minipage}
\end{figure}
\section{Experiments}\label{sec:experiments}
We evaluate VI-SimSiam for two aspects-performance and method of uncertainty prediction.
First, we compare the performance of VI-SimSiam and SimSiam.
We perform a linear evaluation of these methods on ImageNet100~\cite{deng2009imagenet}\footnote{ImageNet100 is a 100-category subset of ImageNet~\cite{deng2009imagenet}.}.
%
Second, we investigate representation uncertainty.
We qualitatively evaluate representation uncertainty by comparing input images and the predicted uncertainty parameter $\kappa$.
Then, we examine the relationship between uncertainty and classification accuracy.
We also study how DINO predicts representation uncertainty.
\subsection{Linear evaluation}\label{linear_eval}
We conduct self-supervised pretraining with ImageNet100 dataset without labels to learn image representations using SimSiam and VI-SimSiam at 50, 100, 200, and 500 epochs. 
Then, we trained a linear classifier on frozen representations on the training set of the dataset with the labels. 
Finally, we evaluated it in the test set of the dataset. 
The implementation details are reported in Appx.~\ref{implementation_details}.
We used Top-1 accuracy as an evaluation metric.

Table~\ref{table:result_supervised_imagenet100} shows Top-1 accuracy on the validation split of ImageNet100. 
VI-SimSiam achieves a competitive result to SimSiam in several epochs.
VI-SimSiam significantly outperforms SimSiam, especially when the number of epochs is less, e.g., 50 and 100.
\renewcommand{\arraystretch}{1.2}
\begin{table}[t]
\centering
    \caption{
        Top-1 accuracies of linear evaluation on ImageNet100.
        Each setting is repeated in triplicate to compute mean and standard derivation.
        We report the result as “mean $\pm$ std.”
    }
    \label{table:result_supervised_imagenet100}
    \scalebox{0.75}{
    \begin{tabular}{lrrrr}
          \hline
          Method & 50 epochs & 100 epochs & 200 epochs & 500 epochs \\
          \hline \hline
          SimSiam \cite{chen2021exploring} & \third{41.07}{0.99} & \third{63.86}{1.48} & \best{78.49}{0.89} & \best{81.61}{0.23} \\
          VI-SimSiam & \best{63.89}{2.74} & \best{73.78}{0.58} & \third{76.31}{0.32} & \third{77.49}{0.42} \\
          \hline
    \end{tabular}
    }
\end{table}
\renewcommand{\arraystretch}{1.0}
%
\subsection{Qualitative analysis of uncertainty}\label{sec:qualitative_result}
We evaluate uncertainty estimation qualitatively by comparing an input image to the predicted concentration $\kappa$, a parameter related to the uncertainty.
We use images from the validation split of ImageNet100.
We use the model pretrained with ImageNet100 on 100 epochs.
Fig.~\ref{fig:kappa_with_images} shows the images for which $\kappa$ is in the top $1\%$ and those for which $\kappa$ is in the bottom $1\%$.
When $\kappa$ is low, i.e., the uncertainty of the latent variable is high, there are few noticeable features in the input images. 
This result shows that our method can estimate high uncertainty for semantically uncertain images.
Additional image examples estimated to have high or low uncertainty are shown in Appx.~\ref{sec:another_result}.
%
\subsection{Quantitative analysis of uncertainty}
\begin{figure}[t]
    \centering
    \includegraphics[width=0.9\linewidth]{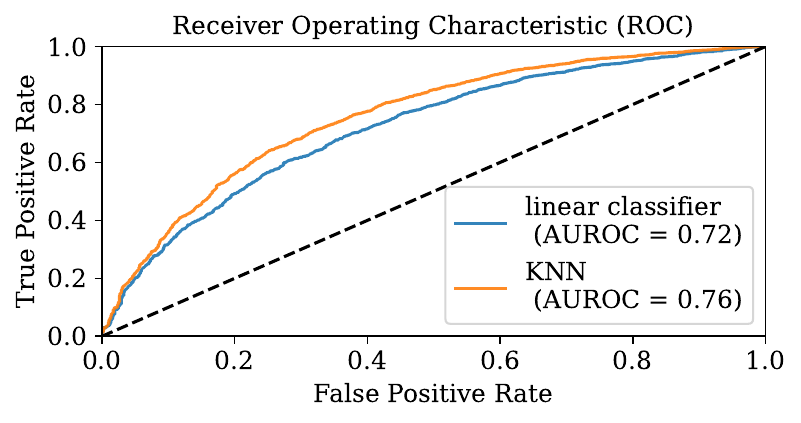}
    \caption{
    ROC and AUROC when the $\kappa$ predicted by VI-SimSiam is used as the confidence score.
    }
    \label{fig:auroc}
\end{figure}

We investigate the effects of uncertainty on a classification task with ImageNet100 dataset.
We evaluate the relationship between uncertainty and Top-1 accuracy using the AUROC (Area Under the Receiver Operating Characteristics Curve) metrics following as per Ref.~\cite{ding2020revisiting}.
Fig.~\ref{fig:auroc} shows ROC and AUROC when the $\kappa$ predicted by VI-SimSiam is used as the confidence score.
We use the VI-SimSiam-trained 500 epochs on ImageNet100.
We use two classifiers -- a linear classifier and a k-nearest-neighbor (KNN) classifier~\cite{wu2018unsupervised}.
The AUROC score of 0.72 by the linear classifier is greater than the chance rate of 0.5 $(p < 0.01)$\footnote{How to calculate the p-value is mentioned in Appx.~\ref{cal_pvalue}.}.
Furthermore, the AUROC score of KNN is 0.76.
The score of the linear classifier is lower than that of the KNN, possibly due to the presence of epistemic uncertainty in the linear classifier itself.
These results suggest we can estimate the difficulty of classifying an image based on uncertainty-related parameters without training the classification model or having the correct labels.
%
%
%
\subsection{Relationship between DINO and uncertainty}\label{sec:dino_unc}
We consider that DINO can estimate uncertainty because it fully estimates distributions, similar to VI-SimSiam. 
In this section, we discuss how DINO expresses uncertainty.
We assume that the entropy of the latent variable, related to the variability of the distribution such as $\kappa$, has a relationship with uncertainty.

\begin{figure}[t]
    \centering    
    \includegraphics[width=0.9\linewidth]{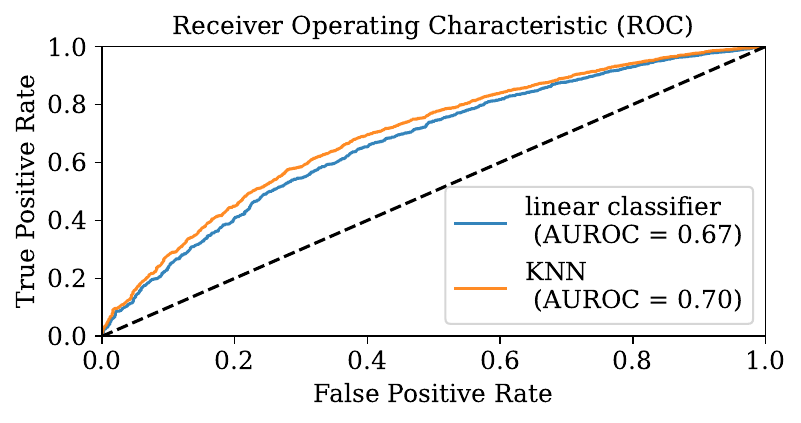}
    \caption{
    ROC and AUROC when the negative entropy predicted by DINO is used as the confidence score.
    }
    \label{fig:entropy_vs_acc}
\end{figure}

To discuss this, we evaluate the relationship between its entropy and Top-1 accuracy using the AUROC metrics.
We use the negative entropy of representation as the confidence score.
The implementation details are reported in Appx.~\ref{implementation_details}.
We use ImageNet100 dataset, and the number of epochs is set to 200.
The results are shown in Fig~\ref{fig:entropy_vs_acc}.
The AUROC score of 0.67 by linear classification is greater than the chance rate of 0.5 $(p < 0.01)$.

\begin{figure}[t]
    \centering
    \includegraphics[width=0.78\linewidth]{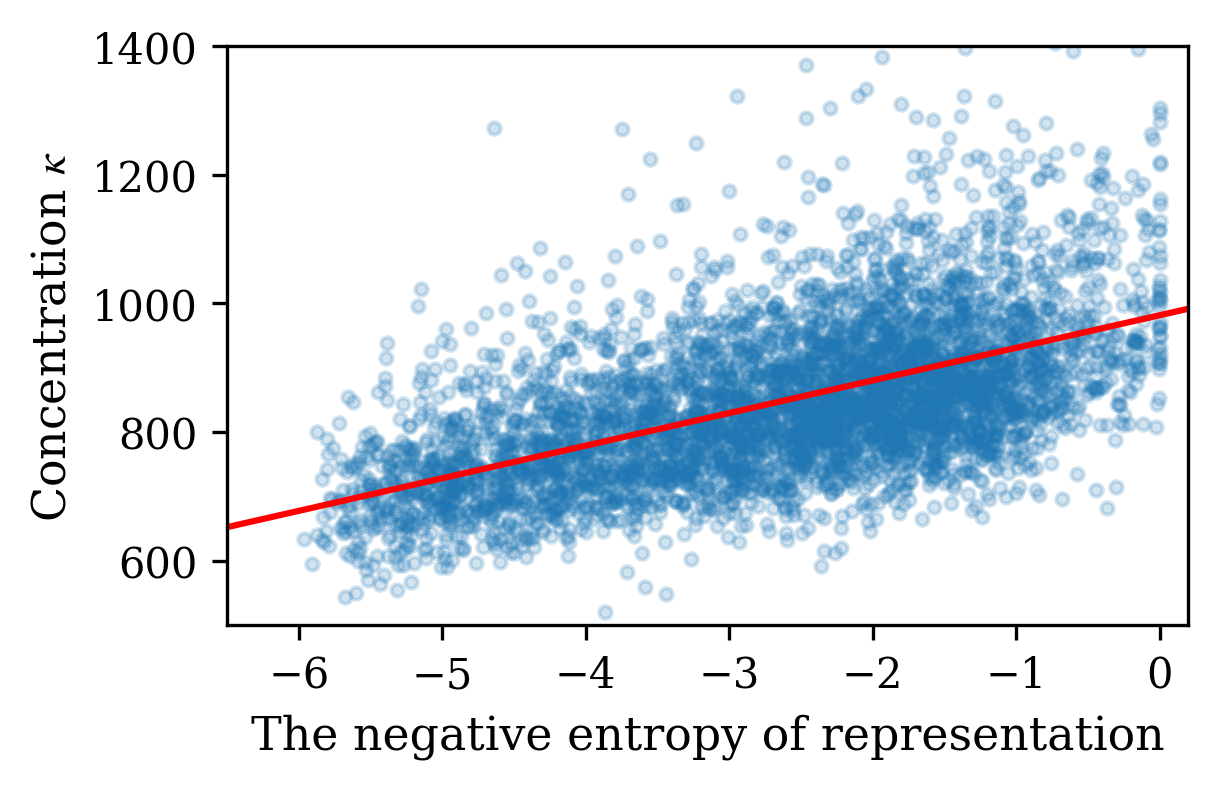}
    \caption{
    Scatter plot of $\kappa$ of VI-SimSiam and entropy of DINO.
    The solid red line is the linear regression line.
    Their correlation coefficient is 0.5459.
    }
    \label{fig:entropy_vs_kappa}
\end{figure}

Fig.~\ref{fig:entropy_vs_kappa} shows the scatter plot of $\kappa$ by VI-SimSiam and the entropy of representation by DINO.
Their correlation coefficient is 0.5459.
The results show that entropy is highly related to $\kappa$ ($p < 0.01$).
Therefore, the entropy of representation by DINO seems to be related to uncertainty.
%
%
\subsection{Relation to image augmentation}
\begin{figure}[t]
    \centering
    \includegraphics[width=0.82\linewidth]{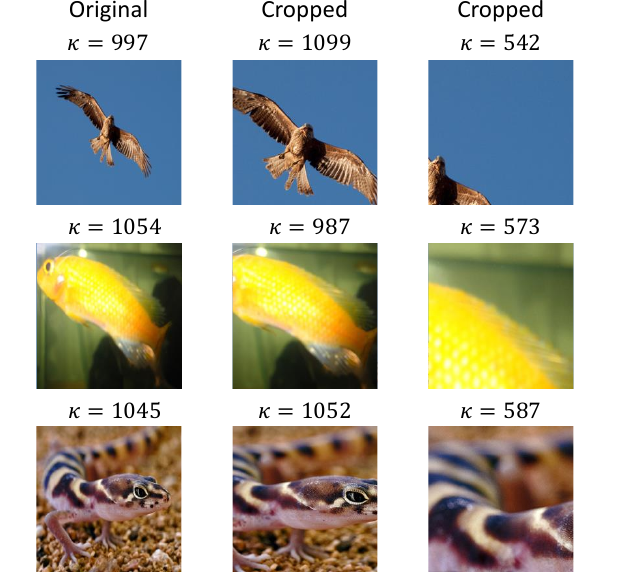}
    \caption{
    A set of cropped images and $\kappa$'s; the images on the left are original. 
    The images in the center and on the right are cropped images.
    This result shows that the $\kappa$ values of images excluding important objects by cropping are low. 
    }
    \label{fig:crop_examples}
\end{figure}

We investigate how image augmentations affect uncertainty. 
We use the validation dataset of ImageNet100. 
We also prepare image augmentations for \textit{blur}, \textit{color jitter}, \textit{grayscale}, and \textit{random crop}.
The random crop scale for \textit{random crop} is set from 0.05 to 1.0.
We perform five augmentations for each image and calculate the average concentration $\kappa$ estimated by the pre-trained model for each augmentation.
The results are shown in Table~\ref{tab:unc_aug_mean}.
The variance of the $\kappa$ of the \textit{random crop} is greater, and its mean value is less than those of other augmentations.
Fig.~\ref{fig:crop_examples} shows examples of images that applied \textit{random crop}.
This result shows $\kappa$ of images excluding important objects by cropping is low. 
Therefore, \textit{random crop} is considered to have a more significant impact on $\kappa$ than the other augmentations.
%
%
\renewcommand{\arraystretch}{1.1}
\begin{table}[t]
\centering
    \caption{
    Mean and standard derivation of $\kappa$ for each augmentation.
    The term "normal" means without augmentations.
    }
    \label{tab:unc_aug_mean}
    \scalebox{0.7}{
    \begin{tabular}{ccccc}
          \hline
           normal & blur & color jitter & grayscale & random crop\\
          \hline \hline
           \third{852.63}{143.69} & \third{836.05}{147.96} & \third{843.94}{144.59} & \third{846.14}{140.53} & \third{812.26}{159.05}\\
          \hline
    \end{tabular}
    }
\end{table}
\renewcommand{\arraystretch}{1.0}
\subsection{Limitation}
\textbf{Performance of Linear evaluation}.
In Sec~\ref{linear_eval}, VI-SimSiam underperforms at 200 and 500 epochs, when compared with SimSiam.
When the cosine similarity is low during training, SimSiam only learns to increase the cosine similarity, while VI-SimSiam learns to also decrease $\kappa$.
It is assumed that VI-SimSiam underperforms at greater epochs because it is not trained to increase cosine similarity for inputs with representations that are difficult to predict.
We compare the cosine similarity of SimSiam and VI-SimSiam in Appx.~\ref{sec:compare_cossim}.
Our method would be more effective if a solution to this issue is proposed.

\textbf{Utilization of Representation Uncertainty}.
Our future work involves proposing novel applications using representation uncertainty.
One example of such use of uncertainty is in the selection of data, the prototype results of which are reported in 
Appx.~\ref{sec:with_kappa}.
%
\section{Conclusion}
In this work, we clarify the theoretical relationship between variational inference and SSL.
Additionally, we propose variational inference SimSiam (VI-SimSiam), which could model the latent variable's uncertainty.
We investigate the estimated uncertainty parameter $\kappa$ from various perspectives.
We derive the relationship between $\kappa$ and input images and between $\kappa$ and classification accuracy.
We also experimentally demonstrate that uncertainty could be estimated even when the latent variable follows the categorical distribution.


{\small
\bibliographystyle{ieee_fullname}
\bibliography{egbib}
}
\newpage
\onecolumn
\appendix
\twocolumn

\section{Proofs}
\subsection{Proof of Proposition \ref{prop:poe}} \label{appx:proof_poe}
Assuming the graphical model in Fig.~\ref{fig:graphical_model} and the non-informative prior,
we can describe the posterior $\posterior \propto \gendist p(z)$ as the PoE of single-modal generative models;
\begin{align}
  \posterior \propto \prod^{M}_{j=1} \gendistj. \label{eqn:proof_poe}
\end{align}
Then, the substitution of the Bayes' theorem relation $\gendistj \propto \posteriorj p(x_{j}|\theta)$ into Eq.~(\ref{eqn:proof_poe}) derives Eq.~(\ref{eqn:poe_posterior}),
where $p(x_{j}|\theta)$ is absorbed into $\eta_{\theta}$.

\subsection{Proof of Claim.~\ref{claim:obj_non_contr}} \label{appx:proof_obj_non_contr}
Substituting Eqs.~(\ref{eqn:varpost}) and (\ref{eqn:vmf_poe}) into Eq.~(\ref{eqn:obj_cross}) yields;
\begin{align}
  & \objcross \propto \sum_{i} \mathbb{E}_{\delta(z-\encodert(x_{i}))}[\eta_{\theta}\prod_{j} \log C_{\mathrm{vMF}}(\kappa) e^{\kappa \encoders(x_{j})^{\mathsf{T}}z}] \nonumber \\
  & \stackrel{+}{\simeq} \sum_{i,j} \mathbb{E}_{\delta(z-\encodert(x_{i}))}[\encoders(x_{j})^{\mathsf{T}}z] = \sum_{i,j} \encoders(x_{j})^{\mathsf{T}} \encodert(x_{i}),
\end{align}
where $\kappa$ and $C_{\mathrm{vMF}}(\kappa)$ are ignored as they are defined to be constants.
%

\subsection{Proof of Claim.~\ref{claim:obj_contr}} \label{appx:proof_obj_contr}
Substituting Eqs.~(\ref{eqn:moe_variational}) and (\ref{eqn:poe_posterior}) into Eqs.~(\ref{eqn:obj_cross}) and (\ref{eqn:obj_uniform}) results in;

\begin{align}
  & \objcross + \objuniform \nonumber \\
  & = \mathbb{E}_{\vardist}\left[ \log \frac{\posterior}{\int \posterior \datadist d\mathbb{X}} \right] \label{eqn:proof_obj_contr1}\\
  & \stackrel{+}{\simeq} \mathbb{E}_{\vardist}\left[ \log \frac{ \prod_{j} \posteriorj }{\int \prod_{j} \posteriorj \datadistj d\mathbb{X} } \right] \label{eqn:proof_obj_contr2}\\
  & = \mathbb{E}_{\vardist}\left[ \log \prod_{j} \frac{ \posteriorj }{\int \posteriorj \datadistj dx_{j} } \right] \\
  & = \sum_{i,j} \mathbb{E}_{\vardisti}\left[ \log \frac{ \posteriorj }{\int \posteriorj \datadistj dx_{j} } \right]. \label{eqn:proof_obj_contr3}
\end{align}
%
By applying Def.~\ref{def:simsiam} and approximating the integral in Eq.~(\ref{eqn:proof_obj_contr3}) by the Monte Carlo method using a mini-batch $\mathcal{B}$,
we can derive Eq.~(\ref{eqn:obj_contr2}).

\section{Implementation details}\label{implementation_details}
The implementation details of networks, pre-training, and linear evaluation are described in this section.
Each experiment needs four NVIDIA V100 GPUs.

\textbf{Experimental settings for networks}; 
The architecture of the encoder $f$ and the predictor $h$ of VI-SimSiam are almost identical to those of DirectPred~\cite{tian2021understanding}. 
We use Resnet18 \cite{he2016deep} as a backbone network of the encoder $f$, unlike the settings of DirectPred. 
We also add the kappa predictor to predict $\kappa$.
The kappa predictor consists of an MLP, which has two fc layers. 
Its input layer has 512 dimensions and its output layer has 1 dimension ($\kappa$).
BN and ReLU activation are applied to its input layer, while soft plus activation is applied to its output layer. 

%
\textbf{Experimental settings for pre-training}; 
We use momentum-SGD for pretraining. The learning rates for VI-SimSiam, SimSiam, and DINO are 0.1, 0.2, and 0.03, respectively. 
The weight decay is 1e-4, and the SGD momentum is 0.9. 
The learning rate has a cosine decay schedule for SimSiam and DINO. 
A learning rate scheduler is not used for VI-SimSiam.
The batch size is 512 for VI-SimSiam and SimSiam.
It is 64 for DINO.
The number of dimensions of the latent variable is 2048.
In DINO, the temperature parameter for the teacher is 0.04, while that for the student is 0.1.
The center momentum rate is 0.9. 
We do not use the momentum encoder as a teacher network in pre-training DINO.

\begin{figure}[t]
    \centering
    \includegraphics[width=0.7\linewidth]{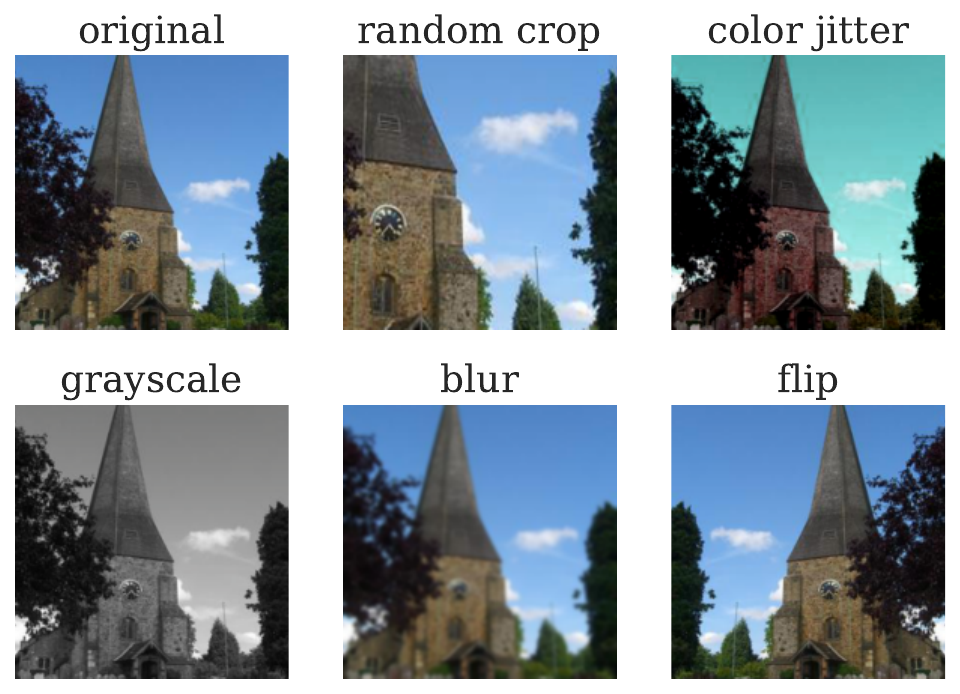}
    \caption{Examples of augmented images.}
    \label{fig:kinds_of_aug}
\end{figure}

We incorporate multi-crop~\cite{caron2020unsupervised} into augmentation.
Multi-crop uses two types of views for training: standard and small-resolution views.
Standard resolution views are generated by the same augmentation setting as that of SimSiam.
SimSiam has five different augmentation types (\textit{blur}, \textit{color jitter}, \textit{flip}, \textit{grayscale} and \textit{random crop}).
Its strength is randomly determined with each applied augmentation.
Fig~\ref{fig:kinds_of_aug} shows examples of augmented views.
Low-resolution views are generated by the augmentation setting of SimSiam with modified random crop and resize parameters.
In the augmentation setting of low-resolution views, the random crop scale is from 0.05 to 0.2, and the size is $96 \times 96$.
We use two standard-resolution views and six low-resolution views per image in each experiment.

\textbf{Experimental settings for linear-evaluation};
We use LARS~\cite{you2017large} as an optimizer for the linear evaluation. 
The learning rate is 1.6. 
The weight decay is 0.0, the SGD momentum is 0.9, and the batch size is 512. 
The image augmentation and preprocessing are referred to as those of SimSiam.
We train 100 epochs and test the model with the highest Top-1 accuracy for the validation set.
When we train it, we crop a random portion of an image and resize it to $224 \times 224$. 
The area of a random cropped image is from 0.08 to 1.0 of the area of the original image.  
When we validate and test it, we resize an image to $256 \times 256$ and crop the center $224 \times 224$. 

ImageNet100 is split into only training and validation.
Using this dataset, we set about 20$\%$ of the training split as a local validation split for tuning parameters. 

\textbf{Experimental settings for k-nearest neighbor classification};
We tuned the parameter $k$ by calculating the top-1 accuracy of the local validation split for 200 $k$s. 
$k$ is determined by Bayesian optimization each time.
%
\section{Additional quantitative analysis of uncertainty}
\begin{figure}[t]
    \centering
    \includegraphics[width=1.0\linewidth]{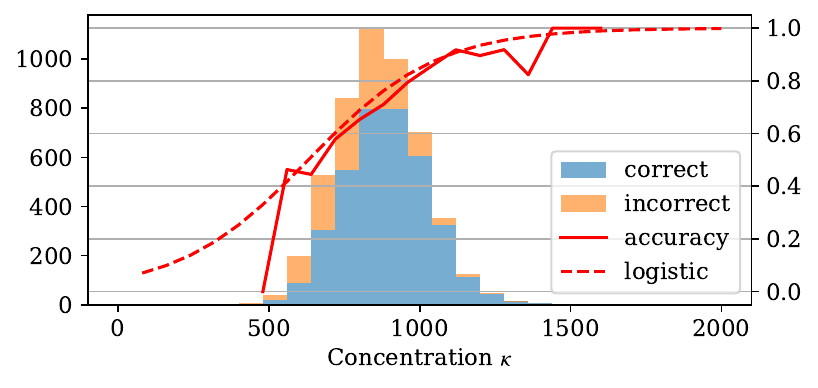}
    \caption{
    Histogram of the estimated concentration $\kappa$.
    The solid red line shows the Top-1 accuracy for each $\kappa$'s range. The dotted line shows the logistic regression curve.
    }
    \label{fig:kappa_vs_acc}
\end{figure}

We investigate the effects of uncertainty on a classification task using methods other than AUROC.
Fig.~\ref{fig:kappa_vs_acc} shows $\kappa$ histograms for correct and incorrect samples and plots each kappa range's Top-1 accuracy and the logistic regression curve.
We use $\kappa$'s predicted by the VI-SimSiam trained 500 epochs on ImageNet100 and labels predicted by linear classifiers with pre-trained features trained on 100 epochs.
As the coefficient of $\kappa$ in logistic regression is positive $(p < 0.01)$\footnote{How to calculate the p-value is mentioned in Appx.~\ref{cal_pvalue}.}, the estimated $\kappa$ increases, and thus, the greater the ease of estimating the class of the input image.

For each estimated label correctness, the mean and standard deviation of estimated $\kappa$ is also given in Table~\ref{tab:kappa_with_correct}.
The mean of the $\kappa$ of image features with the incorrect label is lower than those with the correct label.
%
%
\begin{table}[t]
\centering
    \caption{
    Mean and standard derivation of $\kappa$ for correct and incorrect samples.
    }
    \label{tab:kappa_with_correct}
    \scalebox{0.8}{
    \begin{tabular}{cc}
          \hline
           Correct & Incorrect \\
          \hline \hline
           \third{890.52}{140.53} & \third{805.06}{131.20} \\
          \hline
    \end{tabular}
    }
\end{table}
%
\section{Evaluation of uncertainty estimation of DINO}
\begin{figure}[t]
    \centering
    \includegraphics[width=0.99\linewidth]{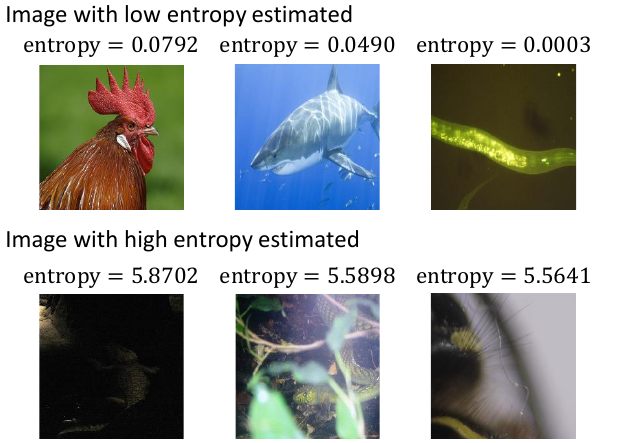}
    \caption{
    \textbf{Input images and the entropy of the latent variable.}
    The greater the entropy, the higher the uncertainty.
    The images with high estimated entropy, i.e., high uncertainty in the representations, appear to have less salient features than the others.
    }
    \label{fig:entropy_with_images}
\end{figure}

We report that the entropy of the latent variable is related to the accuracy of the linear evaluation in Sec~\ref{sec:dino_unc}.
In the current section, we qualitatively evaluate the uncertainty estimation of DINO by comparing entropies and images.
Fig.~\ref{fig:entropy_with_images} shows the images for which entropy is in the top and bottom $1\%$.
When entropy is high, i.e., the uncertainty of the latent variable is predicted to be high, it is difficult to understand the features in the image.
This result shows that DINO can learn the uncertainty of the latent variable, the same as VI-SimSiam.
\section{Linear evaluation with $\kappa$}\label{sec:with_kappa}
We attempt a new linear evaluation using the uncertainty parameter $\kappa$,
which is related to the accuracy of linear evaluation.
Therefore, we use the image with the highest $\kappa$ from several randomly augmented images when testing.
We use 30 augmented images.
A random crop is used for augmentation.
The Top-1 accuracy and the average of $\kappa$ in the model for each epoch are shown in Table~\ref{table:linear_eval_wkappa}.
This result shows a slight improvement in Top-1 accuracy.
Although there is no significant change in the accuracy of the dataset used in this study, it can be effective in datasets where uncertainty can be reduced by image processing, such as noisy images.
\renewcommand{\arraystretch}{1.1}
\begin{table}[t]
\centering
    \caption{
        Top-1 accuracy and the average of $\kappa$ Linear evaluation using the uncertainty parameter $\kappa$.
        ``w/o $\kappa$'' is a normal linear evaluation.
        ``w $\kappa$'' is a linear evaluation with $\kappa$ which uses the image that predicted the highest $\kappa$ from 30 augmented images in the test.
        When we evaluate the ``w $\kappa$'' setting, we test in triplicate and calculate their mean and standard deviation.
    }
    \label{table:linear_eval_wkappa}
    \scalebox{0.8}{
    \begin{tabular}{lrrr}
          \hline
          Method & 100 epochs & 200 epochs & 500 epochs \\
          \hline \hline
          \multicolumn{4}{c}{Top-1 accuracy} \\
          \hline
          w/o $\kappa$ & 73.32 & 76.74 & 77.48 \\
          w $\kappa$ & \third{73.51}{0.43} & \third{77.36}{0.04} & \third{77.61}{0.17} \\
          \hline
          \multicolumn{4}{c}{Average of $\kappa$} \\
          \hline
          w/o $\kappa$ & 867.71& 811.47 & 848.23 \\
          w $\kappa$ & \third{917.81}{0.36} & \third{852.15}{0.42} & \third{893.12}{0.24} \\
          \hline
    \end{tabular}
    }
\end{table}
\renewcommand{\arraystretch}{1.0}
\section{Linear evaluation on Cifar10 dataset}
We conduct self-supervised pre-training with Cifar10~\cite{krizhevsky2009learning} dataset without labels to learn image representations using SimSiam and VI-SimSiam at 100, 200, and 500 epochs. 
Top-1 accuracy is the evaluation metric.
Table~\ref{table:cifar10} shows Top-1 accuracy on the validation split of ImageNet100. 
VI-SimSiam achieves a competitive result to SimSiam in all epochs.
%
\renewcommand{\arraystretch}{1.1}
\begin{table}[t]
    \centering
    \caption{Top-1 accuracies of linear evaluation on Cifar10.}
    \vspace*{0.1cm}
    \scalebox{0.7}{
    \begin{tabular}{l|rrrr}
          \hline
          Method & 100 epochs & 200 epochs & 500 epochs \\
          \hline \hline
          SimSiam & 77.43 & 85.95 & 91.96\\
          VI-SimSiam & \textbf{89.92} & \textbf{93.02} & \textbf{92.75} \\
          \hline
    \end{tabular}
    }
    \label{table:cifar10}
\end{table}
\renewcommand{\arraystretch}{1.0}
\section{Transfer learning}
We evaluate the effectiveness of the representations for transfer learning on some image classification datasets (Flowers~\cite{nilsback2006visual}, Food~\cite{bossard14}, DTD~\cite{cimpoi14describing}, Aircraft~\cite{maji2013fine}, SUN397~\cite{xiao2010sun, xiao2016sun}, Pet~\cite{parkhi2012cats}, and Cars~\cite{KrauseStarkDengFei-Fei_3DRR2013}).
We use representations by models pre-trained at 100 epochs on ImageNet100 dataset.
The implementation details are presented in Appx.~\ref{implementation_details}. 
We report the Top-1 and Top-5 accuracy of VI-SimSiam and SimSiam in Table~\ref{table:result}.
Our method outperforms SimSiam on all datasets at 100 epochs and on some datasets at 200 and 500 epochs.
%
\renewcommand{\arraystretch}{1.1}
\begin{table}[t]
    \centering
    \caption{
    Linear classification Top-1 accuracy in transfer learning. 
    We use pre-trained 100 epochs of representations on ImageNet100 for each method.
    For the 200 and 500 epochs, the results of one experiment are presented, and for the 100 epochs, the mean and standard deviation of three experiments are given.
    }
    \label{table:result}
    \scalebox{0.6}{
    \begin{tabular}{l|rrrrrrrr}
          \hline
          Method & epoch & {Flowers} & {Food} & {DTD} & {Aircraft} & {SUN397} & {Pet} & {Cars} \\
          \hline \hline
          SimSiam \cite{chen2021exploring} & 100 & \third{14.8}{0.5}  & \third{46.7}{1.6} & \third{35.7}{7.1} & \third{10.4}{1.3} & \third{39.5}{1.8} & \third{40.5}{2.5} &\third{8.5}{1.5} \\
          VI-SimSiam & 100 & \best{64.5}{0.5} & \best{48.4}{0.6} & \best{51.8}{0.5} & \best{16.2}{1.3} & \best{46.4}{0.3} & \best{50.0}{0.9} & \best{13.7}{0.4} \\
          \hline
          SimSiam & 200 & 15.6 & \underline{\textbf{55.8}} & \underline{\textbf{55.8}} & 16.0 & \underline{\textbf{51.2}} & \underline{\textbf{54.5}} & \underline{\textbf{16.2}}   \\
          VI-SimSiam & 200 & \underline{\textbf{65.3}} & 49.4 & 53.7 & \underline{\textbf{17.4}} & 47.4 & 50.8 & 14.7 \\
          \hline
          SimSiam & 500 & 28.0 & \underline{\textbf{59.2}} & \underline{\textbf{58.4}} & 17.7 & \underline{\textbf{54.0}} & \underline{\textbf{56.9}} & \underline{\textbf{18.6}} \\
          VI-SimSiam & 500 & \underline{\textbf{64.8}} & 46.1 & 51.7 & \underline{\textbf{17.9}} & 47.9 & 50.9 & 15.4 \\
          \hline
    \end{tabular}
    }
\end{table}
\renewcommand{\arraystretch}{1.0}
\section{Semi-supervised learning}
We evaluate the performance of the proposed method in a more realistic experimental setting, semi-supervised learning.
Considering the excessive cost of labeling and the situation where only some data are labeled, we use labels for only 1\% or 10\% of the total ImageNet100.
After pretraining networks without labels at 100 epochs, we fine-tune the entire networks at 200 epochs with labeled data, for a batch size of 128.
Table \ref{table:result_semisupervised} presents the Top-1 accuracy of each setting. 
Herein, VI-SimSiam consistently outperformed SimSiam for 1\% of labels.
\renewcommand{\arraystretch}{1.1}
\begin{table}[t]
\centering
    \caption{
        Top-1 accuracy under semi-supervised learning on ImageNet100.
        We pretrain models on ImageNet100 without labels for each method. 
        Then, we fine-tune them on 1 \% or 10 \% of ImageNet100 with labels.
        For the 200 and 500 epochs, the results of one experiment are presented, and for the 100 epochs, the mean and standard deviation of three experiments are given.
    }
    \label{table:result_semisupervised}
    \scalebox{0.8}{
    \begin{tabular}{l|rrr}
          \hline
          Method & epoch & 1\% & 10\% \\
          \hline \hline
          scratch & 100 & $10.8\pm0.4$ & $36.8\pm0.6$ \\
          \hline
          SimSiam \cite{chen2021exploring} & 100 & $33.8\pm1.5$ & $64.6\pm0.5$ \\
          VI-SimSiam & 100 & $\mathbf{53.5\pm0.5}$ & $\mathbf{68.8\pm0.0}$ \\
          \hline
          SimSiam & 200 & 49.7 & \textbf{71.2} \\
          VI-SimSiam & 200 & \textbf{56.2} & 70.9 \\
          \hline
          SimSiam & 500 & 42.3 & \textbf{73.3} \\
          VI-SimSiam & 500 & \textbf{58.4} & 72.0 \\
          \hline
    \end{tabular}
    }
\end{table}
\renewcommand{\arraystretch}{1.0}
\section{Linear evaluation of DINO}
We demonstrate a linear evaluation of DINO.
We use ImageNet100 dataset and set the number of epochs to 200. 
We set a batch size to 64 for DINO, unlike SimSiam and VI-SimSiam.
Table~\ref{table:dino_lineval} presents Top-1 accuracy with SimSiam, VI-SimSiam, and DINO.
\renewcommand{\arraystretch}{1.1}
\begin{table}[t]
\centering
    \caption{
        Top-1 accuracy of linear evaluation.
        For all methods, we pretrain a model of 200 epochs.
        We set a batch size to 512 for Simsiam and VI-SimSiam and 64 for DINO.
    }
    \label{table:dino_lineval}
    \scalebox{0.8}{
    \begin{tabular}{lrrr}
          \hline
          & SimSiam & VI-SimSiam & DINO \\
          \hline \hline
          Top-1 accuracy & 78.49 & 76.31 & 76.30\\
          \hline
    \end{tabular}
    }
\end{table}
\renewcommand{\arraystretch}{1.0}
%

%
\section{Test of the relationship between two variables}\label{cal_pvalue}
This section describes whether a variable $x$ is related to a variable $y$ by testing the score $a$. 
Scores are assumed to be a relationship between two variables, e.g., coefficients from linear or logistic regression, and AUROC scores, among others.
%

First, a score is obtained from $X = \{x_1, x_2, ..., x_N\}$ and $Y = \{y_1, y_2, ..., y_N\}$.
We then set the null hypothesis to $H_0$ and the alternative hypothesis to $H_1$.
Set $H_0$ to mean that $X$ and $Y$ are unrelated and $H_1$ to mean that they are related.
For example, if you want to test whether there is a positive proportion, set $H_0: a \leq 0$, $H_0: a > 0$.
Then, $Y^{\prime}_j = \{y^{\prime}_{j, 1}, y^{\prime}_{j, 2},..., y^{\prime}_{j, N}\}$ is obtained by random shuffling $Y$.
This process creates a set of $M$ dummy objective variables $Y^{\prime}_1, Y^{\prime}_2, ..., Y^{\prime}_M$.
$M$ dummy scores $a^{\prime}_1, a^{\prime}_2, ..., a^{\prime}_M$ are obtained from each $Y^{\prime}_j$ and $X$.
We decide whether to reject $H_0$ or not by regarding $M$ dummy scores as a test statistic.

\section{Comparison of cosine similarity}\label{sec:compare_cossim}
\begin{figure}[t]
    \centering
    \includegraphics[width=0.95\linewidth]{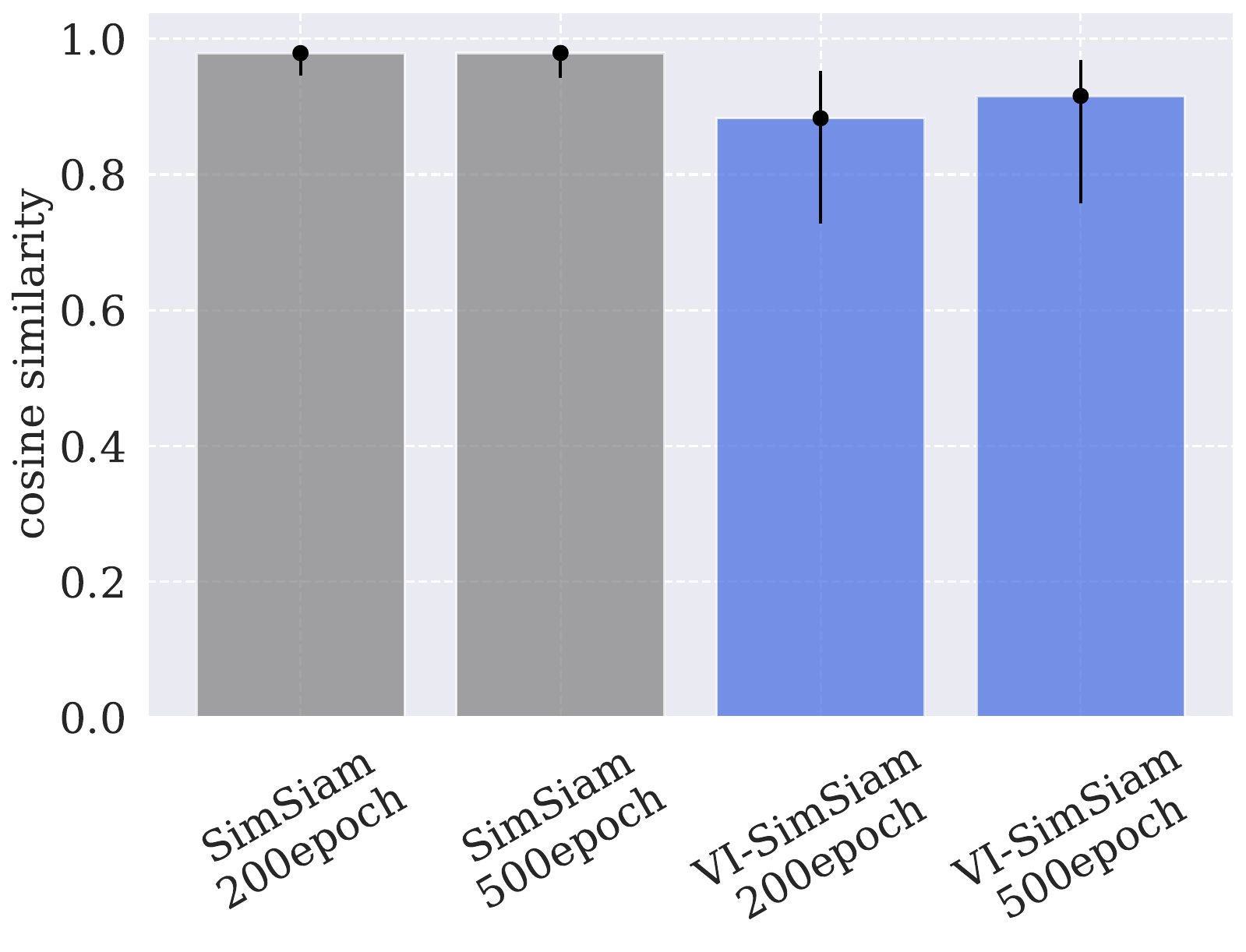}
    \caption{
    Variation of cosine similarity medians of 100000 random augmented image pairs.
    Error bars indicate the upper and lower quartiles.
    }
    \label{fig:cosine_similarity}
\end{figure}

In this section, we discuss the performance of the linear evaluation by comparing cosine similarity.
When the cosine similarity is low during training, SimSiam only learns to increase the cosine similarity, while VI-SimSiam learns to decrease $\kappa$.
The median of cosine similarity of 100000 random augmented image pairs is shown in Fig~\ref{fig:cosine_similarity}.
Thus, the cosine similarity of VI-SimSiam is lower than that of SimSiam.
In some cases, it is assumed that VI-SimSiam learns to minimize the loss by reducing $\kappa$ rather than increasing the cosine similarity for inputs with representations that are difficult to predict.
\section{Additional Experimental Results}\label{sec:another_result}
\begin{figure*}[t]
    \centering
    \includegraphics[width=0.85\linewidth]{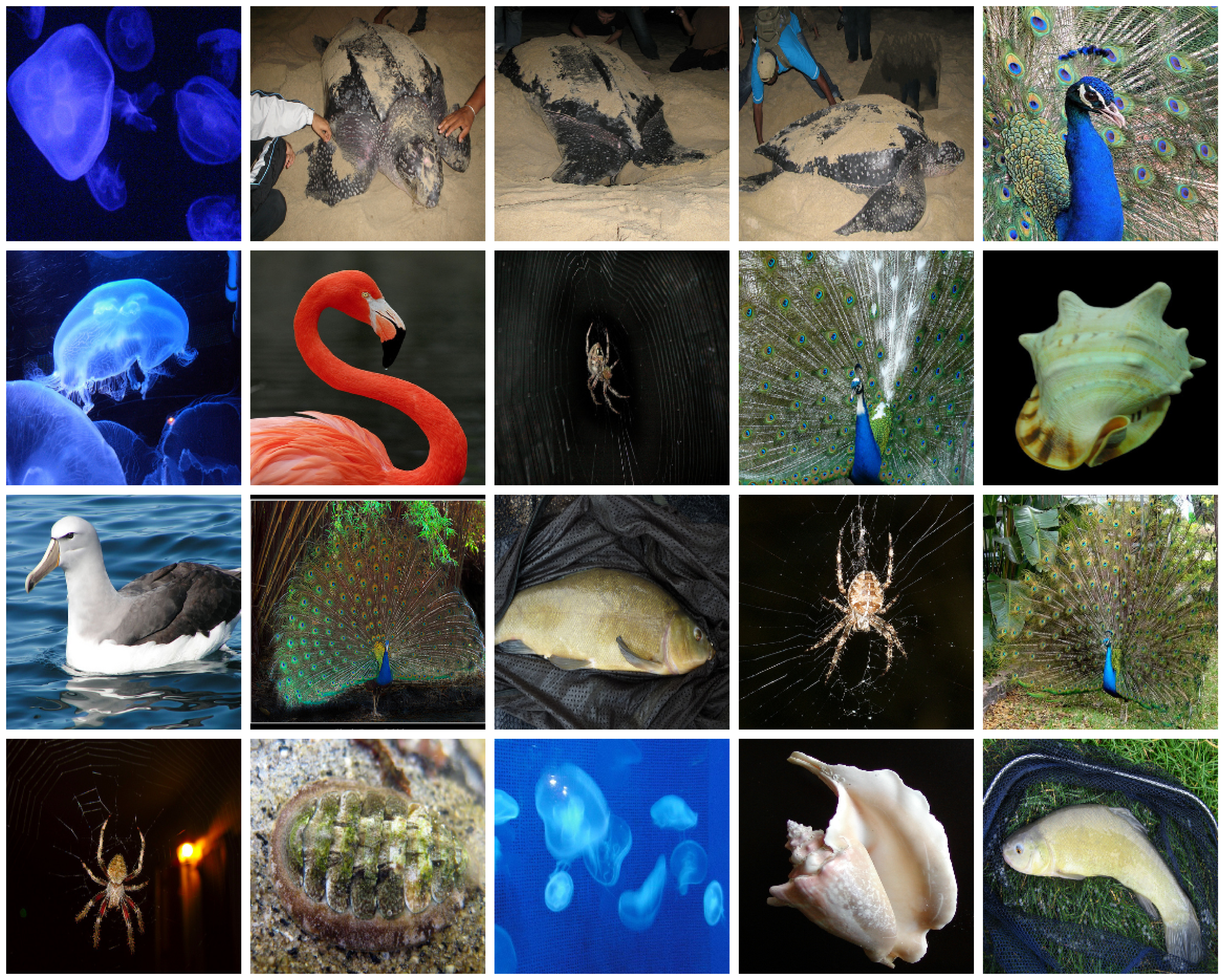}
    \caption{
    Twenty images estimated to have the highest $\kappa$.
    }
    \label{fig:top20_kappa}
\end{figure*}

\begin{figure*}[t]
    \centering
    \includegraphics[width=0.85\linewidth]{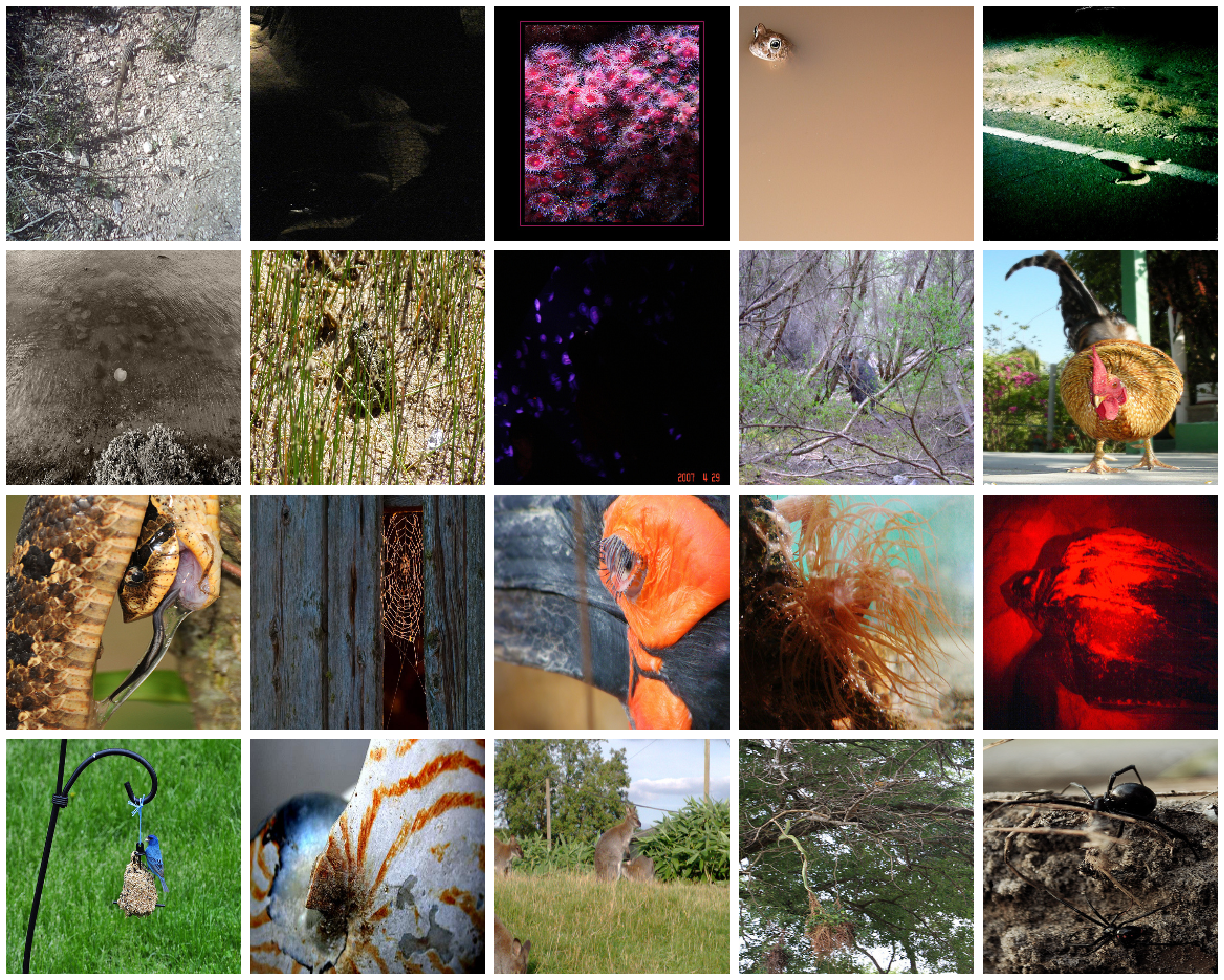}
    \caption{
    Twenty images estimated to have the lowest $\kappa$.
    }
    \label{fig:worst20_kappa}
\end{figure*}

In this section, we show another result of \S~\ref{sec:qualitative_result}.
Fig.~\ref{fig:top20_kappa} shows twenty images estimated to have the highest $\kappa$, i.e., the lowest uncertainty.
On the other hand, Fig.~\ref{fig:worst20_kappa} shows twenty images estimated to have the lowest $\kappa$, i.e., the highest uncertainty.
%

\end{document}